\documentclass[runningheads]{llncs}
\usepackage{graphicx}

\usepackage{tikz}
\usepackage{comment}
\usepackage{amsmath,amssymb} 
\usepackage{color}
\usepackage{rotating}
\usepackage{floatrow}
\newfloatcommand{capbtabbox}{table}[][\FBwidth]

\usepackage[accsupp]{axessibility}  

\begin{document}
\pagestyle{headings}
\mainmatter
\def\ECCVSubNumber{13}  
\def \eg {\textit{e.g.,~}}
\def \ie {\textit{i.e.,~}}
\def \etal {\textit{et al.~}}
\def \fullcircle {360$^{\circ}$~}

\title{Photo-realistic \fullcircle Head Avatars in the Wild}

\titlerunning{Photo-realistic \fullcircle Head Avatars in the Wild}
%
\author{Stanislaw Szymanowicz\inst{1,2}\thanks{Work done while at Microsoft.} \and
Virginia Estellers\inst{1} \and
Tadas Baltru\v{s}aitis\inst{1} \and \\
Matthew Johnson\inst{1}
}

\authorrunning{S. Szymanowicz et al.}
%
\institute{Microsoft \and University of Oxford  \\
\email{stan@robots.ox.ac.uk, \{virginia.estellers,tadas.baltrusaitis,matjoh\}@microsoft.com}}
\maketitle

\vspace{-0.7cm}
\begin{abstract}
Delivering immersive, 3D experiences for human communication requires a method to obtain \fullcircle photo-realistic avatars of humans. To make these experiences accessible to all, only commodity hardware, like mobile phone cameras, should be necessary to capture the data needed for avatar creation. For avatars to be rendered realistically from any viewpoint, we require training images and camera poses from all angles. However, we cannot rely on there being trackable features in the foreground or background of all images for use in estimating poses, especially from the side or back of the head. To overcome this, we propose a novel landmark detector trained on synthetic data to estimate camera poses from \fullcircle mobile phone videos of a human head for use in a multi-stage optimization process which creates a photo-realistic avatar. We perform validation experiments with synthetic data and showcase our method on \fullcircle avatars trained from mobile phone videos.
\end{abstract}

\vspace{-0.7cm}
\section{Introduction}

Immersive interaction scenarios on Mixed Reality devices 
require rendering human avatars from all angles. To avoid the uncanny valley effect, these avatars must have faces that are photo-realistic. It is likely that in the future virtual spaces will become a ubiquitous part of every-day life, impacting everything from a friendly gathering to obtaining a bank loan. For this reason we believe high-quality, \fullcircle avatars should be affordable and accessible to all: created from images captured by commodity hardware, \eg from a handheld mobile phone, without restrictions on the surrounding environment.

\begin{figure}
    \centering
    \includegraphics[width=\columnwidth]{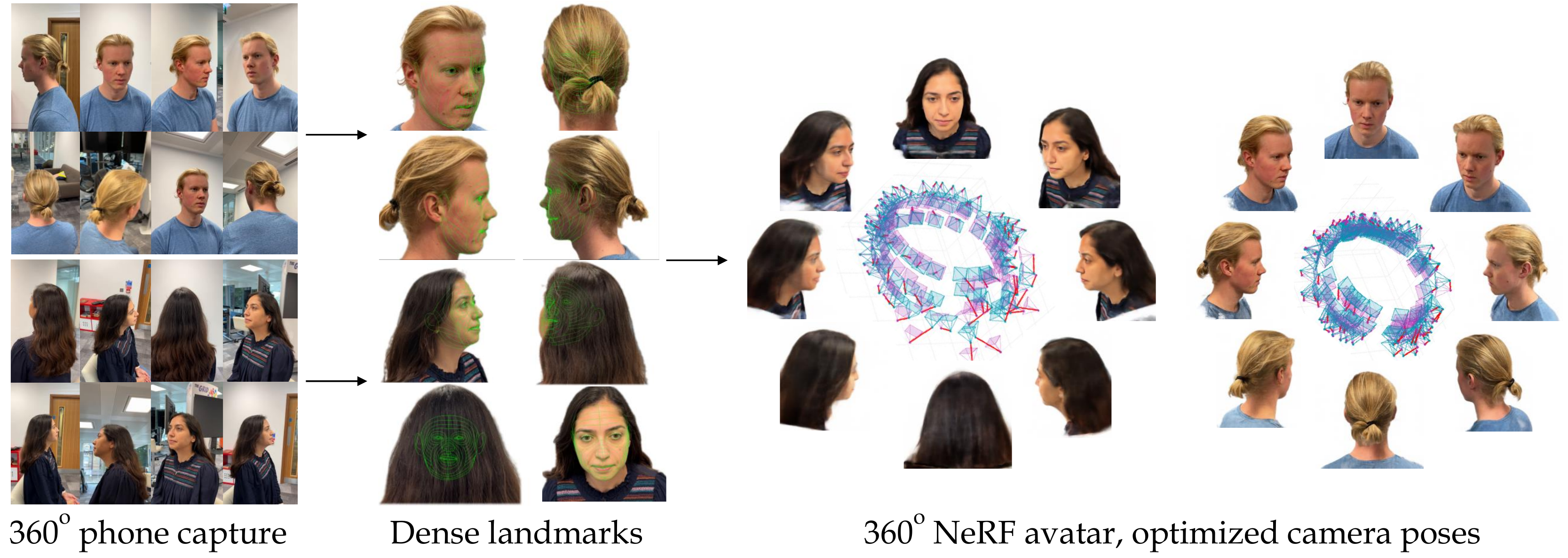}
    \caption{Our system creates photo-realistic \fullcircle avatars from captures from a mobile phone capture and without constraints on the environment. Cameras are registered 
    from full \fullcircle pose variation, and our multi-stage optimization pipeline allows for high quality avatars.}
    \label{fig:teaser}
\end{figure}

Obtaining data to train a \fullcircle photo-realistic avatar `in the wild' is challenging due to the potential difficulty of camera registration: traditional Structure-from-Motion pipelines rely on reliable feature matches of static objects across different images. Prior work limits the captures to a 120$^\circ$ frontal angle which allows the use of textured planar objects that are amenable to traditional feature detectors and descriptors (\eg a book, markers, detailed wall decoration). However, in many \fullcircle captures from a mobile phone in unconstrained environments neither the background nor the foreground can be depended upon to provide a source of such matches.

There are several properties of \fullcircle captures in the wild which pose serious challenges to camera registration and avatar model learning. First, the space being captured is likely to either have plain backgrounds (\eg white walls) and/or portions of the capture in which the background is an open space, leading to defocus blur and the inclusion of extraneous, potentially mobile objects (\eg pets, cars, other people). Second, in order to obtain the needed details on the face and hair the foreground subject will likely occupy much of the frame. While the face can provide some useful features for camera registration, its non-planar nature combined with changes in appearance due to lighting effects make it less than ideal. Further, while the back of the head can produce many features for tracking the matching can become highly ambiguous due to issues with hair, \ie specular effects and repeated texture.

To address the challenges of \fullcircle captures we propose a multi-stage pipeline to create 3D photo-realistic avatars from a mobile phone camera video. 
We propose using head landmarks to estimate the camera pose. 
However, as most facial landmark detectors are not reliable at oblique or backward-facing angles, we propose using synthetic data to train landmark detectors capable of working in full \fullcircle range. 
We use the predicted landmarks to provide initialization of the 6DoF camera poses, for a system which jointly optimizes a simplified Neural Radiance Field with the camera poses. Finally, we use the optimized camera poses to train a high quality, photo-realistic NeRF of the subject.

The contributions of our work are three-fold: (1) a reliable system for camera registration which only requires the presence of a human head in each photo, (2) a demonstration of how to leverage synthetic data in a novel manner to obtain a DNN capable of predicting landmark locations from all angles, and (3) a multi-stage optimization pipeline which builds \fullcircle photo-realistic avatars with high-frequency visual details using images obtained from a handheld mobile phone `in the wild'.

\section{Related work}

\textbf{Photo-realistic avatars in the wild.} Recent works on surface\cite{chen2022authentic_volumetric,grassal2021neural,zheng2022imavatar} and volumetric avatars of human heads\cite{nerface,park2021nerfies} enroll the avatar from a dynamic video of the subject, and they constrain the capture to only take frontal views, either through limitations of the face tracker\cite{chen2022authentic_volumetric,nerface,grassal2021neural,zheng2022imavatar} or by requiring a highly textured, static background\cite{park2021nerfies} to acquire camera poses. Instead, our subject is static and captured by another person. In contrast to all of the approaches above, this scenario enables photo-realistic novel-view renders from the full \fullcircle range, whereas previous works have only shown results from within a frontal 180$^\circ$.

\noindent \textbf{Detecting occluded landmarks.} Landmark detection \cite{wood2022dense} is a challenging task even under $\pm 60^{\circ}$ yaw angle variation \cite{Zhu2016}. Existing datasets and approaches are limited to $\pm 90^{\circ}$\cite{deng2019menpo,Zhu2016}, partially due to the difficulty of reliably annotating dense landmarks in such data. Fortunately, recent work on using synthetic data in face understanding\cite{wood2021fake} has shown evidence of being able to detect landmarks even under severe occlusions. We push this capability to the limit: we aim to detect all landmarks even if completely occluded, \ie observing the back of the head. Unlike prior work, we use a synthetic dataset with full \fullcircle camera angle variations. The synthetic nature of the dataset allows us to calculate ground truth landmark annotations which could not be reliably labelled on real data.

\noindent\textbf{Camera poses.} We estimate initial camera poses from detected face landmarks, similarly to object pose detection from keypoints\cite{Lowe99Sift}. We are not constrained by the requirement of highly textured planar scene elements which Structure-from-Motion (SfM) methods impose. We are therefore able to capture our training data in the wild: we perform robust camera registration when a head is observed in the image, while SfM approaches can fail to register all cameras when insufficient matches are available, such as when observing dense dark hair.

Optimizing camera poses jointly with NeRF has been the topic of BARF\cite{lin2021barf} and NeRF-{}-\cite{wang2021nerfmm}. We use the optimization method from BARF, but it only uses one, coarse, NeRF network. Therefore, for maximum visual quality we run a second optimization, starting from the optimized camera poses but initializing the coarse and fine networks from scratch and optimizing them jointly with camera poses. We hence achieve the high-frequency visual detail that the two-stage sampling provides, while optimizing the camera poses to avoid artefacts.

\begin{figure}
    \centering
    \includegraphics[width=\columnwidth]{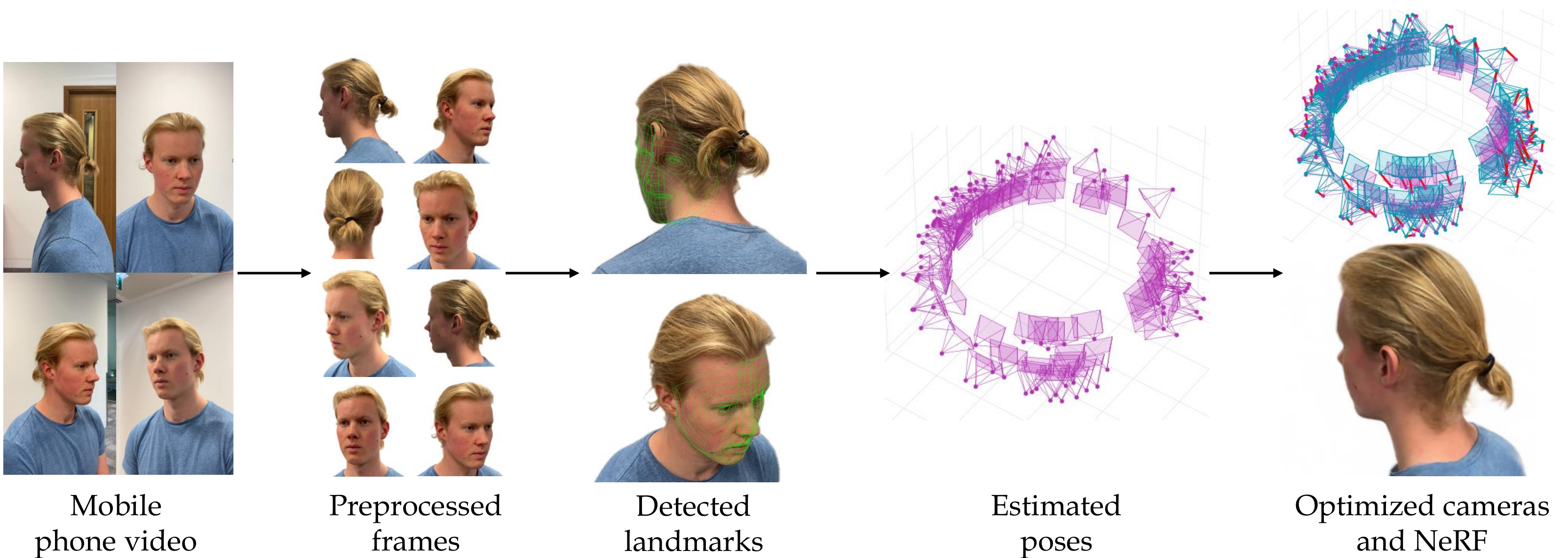}
    \caption{We build a \fullcircle Neural Radiance Field of a person from a video captured from a mobile phone. In each selected  frame we detect dense landmarks, and we use them to register the cameras from all angles. In the joint optimization process, we refine camera poses jointly with NeRF, obtaining a highly detailed volumetric avatar.}
    \label{fig:method}
\end{figure}

\section{Method}

We estimate camera pose by solving a Perspective-n-Point (PnP) problem that matches 3D points in the world and their corresponding 2D projections in the images. The 2D image points are dense face landmarks while the 3D world points are the same landmarks annotated on a template face model\cite{wood2021fake}. We use standard convolutional neural networks (CNN) to estimate landmarks and standard RANSAC PnP solvers to solve for camera pose. Our results are possible because our landmark detector, trained with synthetic data, provides good landmarks for \fullcircle views of a face within $\pm 60 ^{\circ}$  of elevation.

\noindent\textbf{Landmarks for Pose Estimation.} The main difference between our landmark detector and competing approaches is the number and quality of landmarks, specially for non-frontal views. While a human can consistently label frontal face images with tens of landmarks, it is impossible to annotate an image of the back of the head with hundred of landmarks in a manner consistent in 3D. Using synthetic data allows us to generate images with perfectly consistent labels. We use the framework introduced by Wood \etal\cite{wood2021fake} to render a synthetic training dataset of 100k images that contain extreme head and camera pose variations, covering the \fullcircle range. We use it to train an off-the-shelf ResNet-101 landmark detector model to directly regress 703 landmark positions from an image\cite{wood2022dense}.

The camera poses estimated from landmarks might contain errors because the occluded landmarks are a rough estimate based on the priors learned by the CNN, but they are good enough for our avatar generation system.

\noindent\textbf{Avatar Generation.} We choose to represent our 3D avatar with a Neural Radiance Field\cite{nerf} (NeRF): a volumetric representation of the scene because it can generate novel views with a high level of photo-realism and can handle human hair and accessories seamlessly\cite{Vicini2021NonExponential}. This is important for in-the-wild captures where no assumptions of subjects appearance are possible. 

We use the original representation, architecture and implementation details of NeRF\cite{nerf}. We optimize the camera poses jointly with the radiance field in a coarse-to-fine manner, as in BARF\cite{lin2021barf}. While BARF\cite{lin2021barf} uses only the coarse MLP, we improve on its proposed optimization scheme with a second optimization step to achieve higher fidelity. The steps are as follows: (1) Initialize camera poses from landmarks and optimize them jointly with the coarse MLP following BARF\cite{lin2021barf}. Discard the coarse MLP. (2) Initialize camera poses from step 1 and optimize them jointly with both coarse and fine MLPs initialized from scratch.

\section{Data}

We use two types of data: (1) synthetic data, allowing us to perform experiments in a controllable manner and (2) \fullcircle mobile phone captures of real people. 

We generate three \textbf{synthetic} faces. For each, we render 90 400x400 px training and 10 testing images from camera poses varying across the entire \fullcircle range and elevation angle between $-25$ and $45$ degrees, simulating a \fullcircle capture that could be done by someone walking around a person sitting still in a chair.

We collect two \textbf{real} datasets: in each, one person sits still in a chair, and another person takes a video of them from a mobile phone. The capture covers \fullcircle angles in forms of 3 orbits at different elevation levels. The data is preprocessed for training the avatar by selecting 100 images without motion blur, undistorting them, cropping to a square, segmenting out the background and resizing to 400x400 resolution. 

\section{Experiments}

\begin{figure}
    \centering
    \includegraphics[width=\columnwidth]{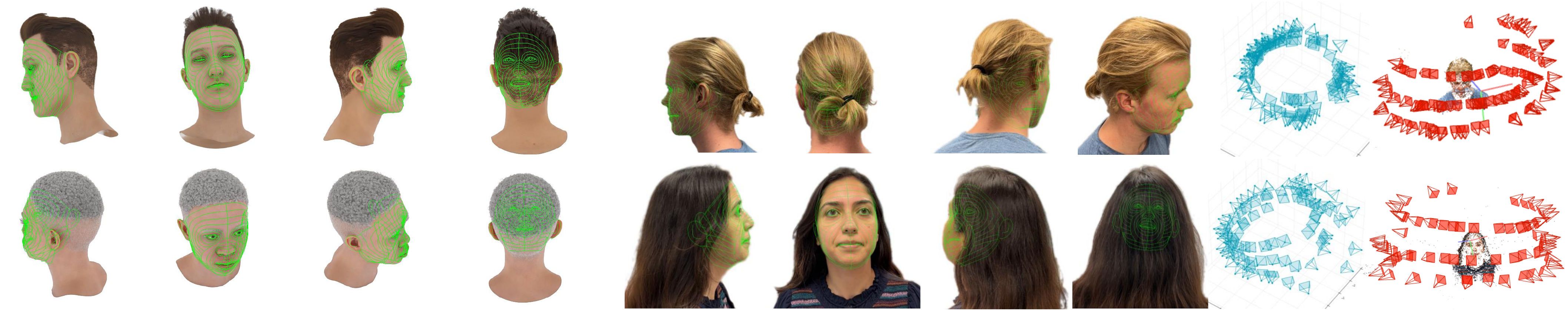}
    \caption{Detected landmarks are plausible on both synthetic and real faces from all viewpoints and allow to register all cameras (blue frustums). In our sequences SfM pipelines like COLMAP only register front-facing cameras (red frustums).} 
    \label{fig:experiments/dense_landmarks}
\end{figure}

\textbf{Dense landmark detection and pose registration.} Fig.~\ref{fig:experiments/dense_landmarks} shows dense landmark detection on real and synthetic images of faces. The results on real images of the back of the head show that a network trained with synthetic data learns correct priors to estimate plausible landmarks from weak cues like ears or the head orientation. With these landmarks, we can register cameras from images of \fullcircle angles around the head (blue frustums in Fig.~\ref{fig:experiments/dense_landmarks}) where traditional SfM pipelines like COLMAP\cite{schoenberger2016sfm,schoenberger2016mvs} (red frustums in Fig.~\ref{fig:experiments/dense_landmarks}) partially fail because images of the back of the head do not have enough reliable matches.

\begin{figure}
    \begin{tabular}{c@{\hspace{0.3cm}}c@{\hspace{0.2cm}}c}
         Landmark + PnP & Optimized & \\
    
        \includegraphics[width=0.27\columnwidth]{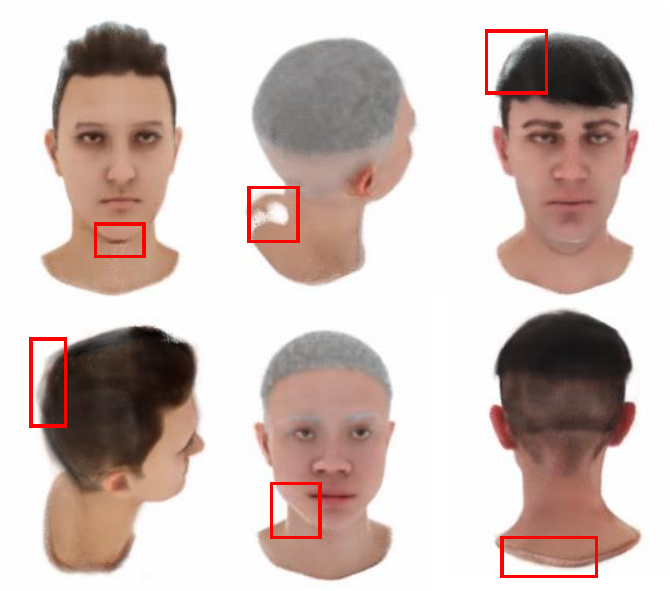} & 
        \includegraphics[width=0.27\columnwidth]{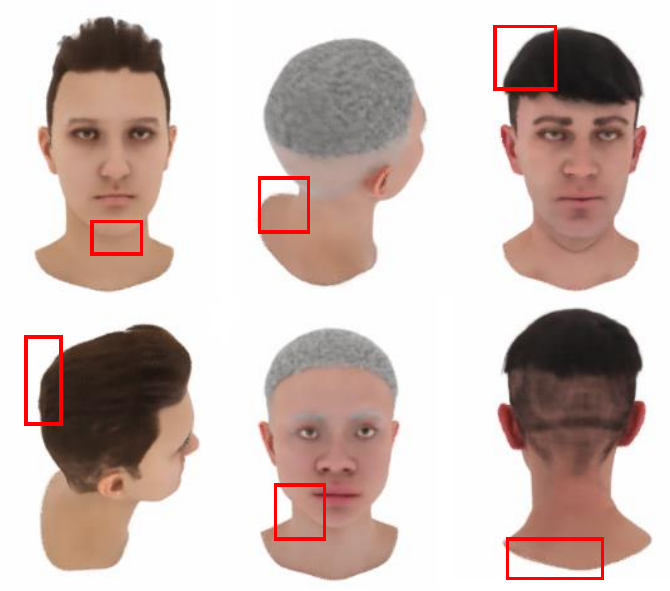}
        &
        \begin{tabular}[b]{c|c|c|c}
            Cameras & ~PSNR~ & ~R(deg)~ & ~t(cm) \\
            \hline
            GT & 35.18 & - & - \\
            PnP & 30.34 & 2.60 & 7.66 \\
            Optim. & 35.19 & 0.334 & 1.66 \\
            \hline
        \end{tabular}
    
    \end{tabular}
    \caption{Novel viewpoints rendered from a NeRF trained with camera poses from PnP exhibit artefacts (left). Optimizing (Optim.) poses jointly with NeRF removes the artefacts (middle), recovers the visual quality of NeRF trained with Ground Truth cameras (GT) and the registration error (right). Best viewed zoomed in.}
    \label{fig:experiments/qualitative_synthetics_360}

\end{figure}
\noindent \textbf{Avatars of synthetic faces.} We compare avatar quality as trained from 1) ground truth poses, 2) poses as estimated from dense landmarks, 3) poses optimized jointly with NeRF when initialized from dense landmarks. Fig.~\ref{fig:experiments/qualitative_synthetics_360} qualitatively compares novel view renders from different approaches and its associated table provides quantitative metrics w.r.t. ground truth camera poses.

Training a NeRF avatar with noisy poses estimated from landmarks leads to artefacts in the renders, but using them as initialization and optimizing jointly with NeRF removes the artefacts and successfully recovers camera poses. The error in camera poses, leads to a significant drop in PSNR (Fig.~\ref{fig:experiments/qualitative_synthetics_360}) of novel-view renders. However, optimizing camera poses jointly with NeRF leads to novel-view quality on par with that obtained using the ground truth cameras. 

\begin{figure}
    \centering
    \includegraphics[width=\columnwidth]{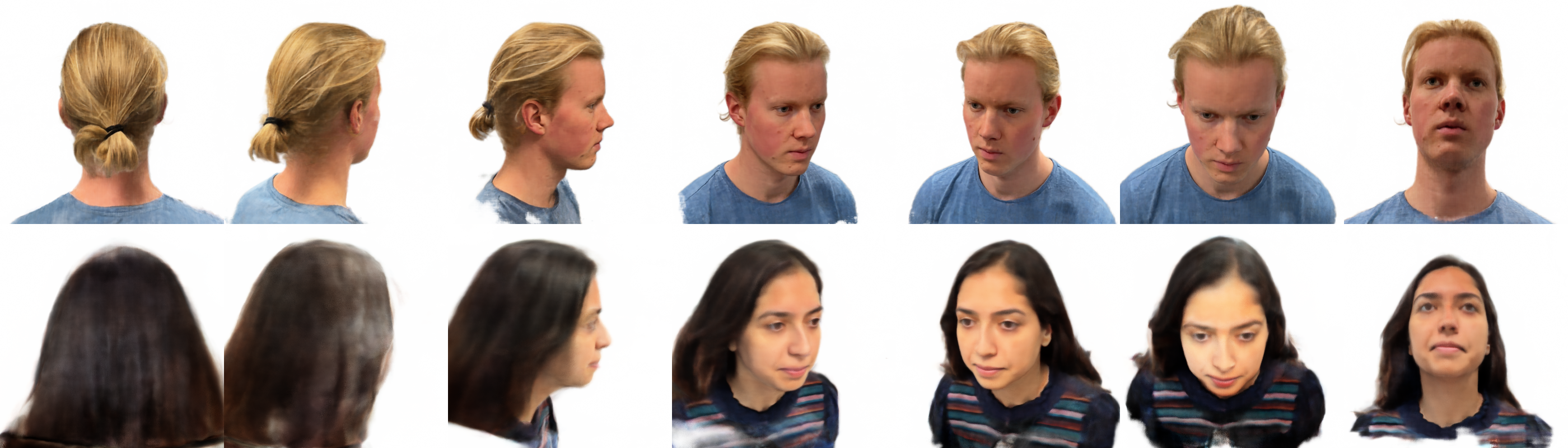}
    \caption{Novel view renders of \fullcircle avatars trained from images captured from a phone.}
    \label{fig:experiments/real_qualitative_360}
\end{figure}

\noindent \textbf{Avatars of real faces.} We test our full pipeline on two real captures. Fig.~\ref{fig:experiments/real_qualitative_360} shows novel-view renders of avatars trained on \fullcircle capture. The detail in renders is visible even from the back, where the initial camera poses were likely with the most error. We hypothesize that white regions on the torso stem from being cropped by segmentation masks in training data.

\begin{figure}
    \centering
    \includegraphics[width=\columnwidth]{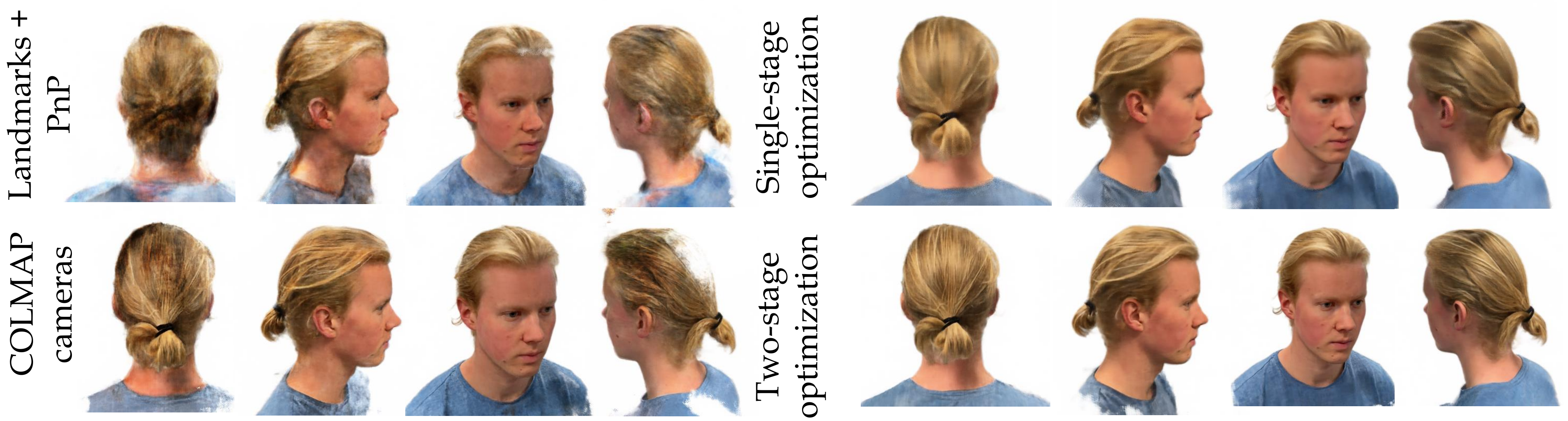}
    \caption{Training a NeRF with cameras with a registration error results in artefacts in renders from novel viewpoints (left). Using BARF\cite{lin2021barf} removes the artefacts, but lacks the high-frequency details (top right). Adding our second optimization stage leads to an avatar with high-frequency detail but without artefacts (bottom right). 
}
    \label{fig:experiments/real_comparison}
\end{figure}

Fig.~\ref{fig:experiments/real_comparison} shows the comparison of our method to the baselines on real data: (1) using camera poses as estimated from the dense landmarks,  (2) using cameras from manually joined models output by off-the-shelf Structure-from-Motion software COLMAP, 
(3) optimizing a coarse NeRF jointly with camera poses initialized with the estimate from dense landmarks, (4) two-stage optimization process where second stage optimizes both fine and coarse NeRF from scratch with poses initialized at the output of (3). 

Error in camera poses result in avatars with artefacts (Fig.~\ref{fig:experiments/real_comparison}, left). NeRF optimized jointly with camera poses in the same manner as in BARF does not exhibit the artefacts but lacks in visual detail due to only training the coarse network (Fig.~\ref{fig:experiments/real_comparison}, top right). Bottom right row of Fig.~\ref{fig:experiments/real_comparison} illustrates that adding the second optimization stage recovers the fine detail while avoiding artefacts, because the second stage starts from a good initialization of camera poses, and optimizes them jointly with both the coarse and fine network.

\section{Conclusions}

\textbf{Limitations and future work.} First, while our current formulation only renders a static avatar, recent work\cite{garbin2022voltemorph} could be used to animate it using geometry-driven deformations. Secondly, we train NeRF on images with background segmented out, and segmentation errors, \eg on the hair, can lead to incorrectly learnt colour which is then revealed from different viewpoints resulting in artefacts. Future work should aim to refine masks in the optimization process by using the projection of the density field. Finally, we observed that the camera pose estimates for the subject with hair covering the ears resulted in larger camera pose error. While in our experiments we observed that the pose optimization converged to a good solution, future work should (1) investigate the basin of convergence of camera poses and (2) test the method on more subjects.

\noindent \textbf{Conclusions} In this paper we demonstrate that we can obtain a \fullcircle avatar from a mobile phone video of a subject gathered `in the wild'. We train a facial landmark DNN on synthetic data which is sampled from the full \fullcircle range of view angles. We then show that the landmarks predicted by this DNN can be used to initialize the camera poses for the mobile phone image sequence. This is the first stage of an optimization pipeline which we demonstrate can progressively fine-tune the camera poses while training a NeRF that can be used to create photo-realistic, highly detailed images of the subject from novel views, including from the back and side. This can be used as the foundation for a \fullcircle avatar of the kind shown in \cite{garbin2022voltemorph}.

\clearpage
%
%
\bibliographystyle{splncs04}
\bibliography{egbib}
\end{document}